# LAME: Layout-Aware Metadata Extraction Approach for Research Articles

**JONGYUN CHOI[1], HYESOO KONG[2], HWAMOOK YOON[2], HEUNG-SEON OH[3], and YUCHUL JUNG[1*]**

[1]Department of Computer Engineering, Kumoh National Institute of Technology (KIT), Gumi, South Korea

[2]Korea Institute of Science and Technology Information (KISTI), South Korea

[3]School of Computer Science and Engineering, Korea University of Technology and Education (KOREATECH), sCheonan, South Korea

**Abstract:** The volume of academic literature, such as academic conference papers and journals, has increased rapidly worldwide, and research on metadata extraction is ongoing. However, high-performing metadata extraction is still challenging due to diverse layout formats according to journal publishers. To accommodate the diversity of the layouts of academic journals, we propose a novel LAyout-aware Metadata Extraction (LAME) framework equipped with the three characteristics (e.g., design of an automatic layout analysis, construction of a large meta-data training set, and construction of Layout-MetaBERT). We designed an automatic layout analysis using PDFMiner. Based on the layout analysis, a large volume of metadata-separated training data, including the title, abstract, author name, author affiliated organization, and keywords, were automatically extracted. Moreover, we constructed Layout-MetaBERT to extract the metadata from academic journals with varying layout formats. The experimental results with Layout-MetaBERT exhibited robust performance (Macro-F1, 93.27%) in metadata extraction for unseen journals with different layout formats.

**Keywords:** Automatic layout analysis, Layout-MetaBERT, Metadata extraction, Research article

## 1 Introduction

With the development of science and technology, the number of related academic papers distributed periodically worldwide has reached more than several hundred thousand. However, their layout styles are as diverse as their subjects and publishers although the portable document format (PDF) is widely used globally as a standardized text-based document provision format. For example, the information order is inconsistent when converting such a document to text because no layout information separating the document content is provided. Thus, extracting meaningful information such as metadata, including title, author names, affiliations, abstract, and keywords, from a document is quite challenging.

Research on extracting metadata or document objects from PDF documents using machine learning has increased [1-7]. In aspects of natural language processing (NLP) approach, Open-source software, such as Content ExtRactor and MINEr (CERMINE) [4] and GeneRation of Bibliographic Data (GROBID) [5], automatically extract metadata using the sequential labeling technique but generally do not take the layouts into account in detail. Therefore, they do not show reasonable metadata extraction performances for every research article due to their diverse (and sometimes bizarre) layout formats.

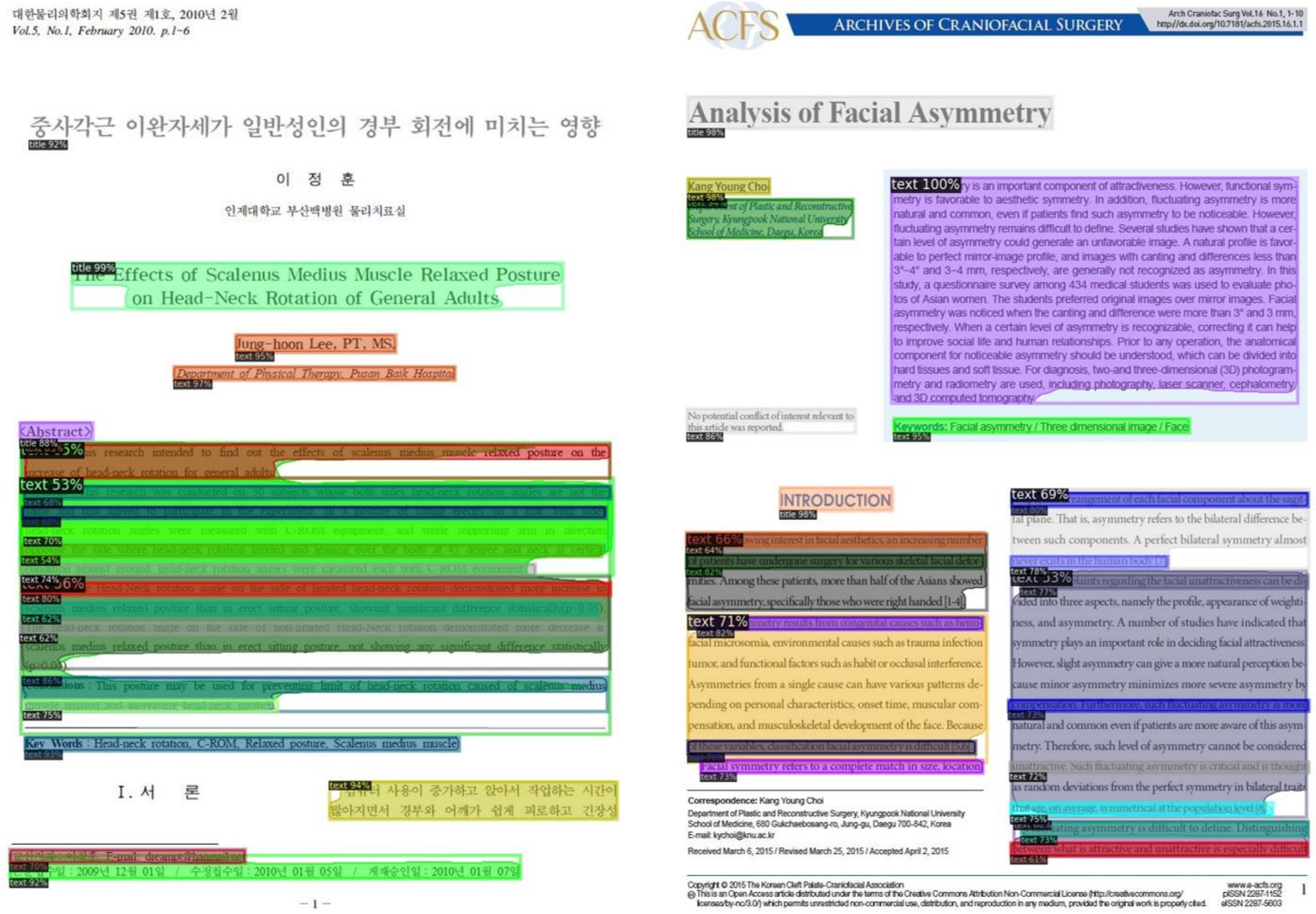

**Figure 1:** Layout analysis results of PubLayNet [1].

Unlike existing NLP based metadata extraction approaches, PubLayNet [1], LayoutLM [7], and DocBank [3] employ object detection models, such as Mask region-based convolutional neural network (Mask R-CNN) [8] and Faster R-CNN [9], to detect the layout of academic literature and extract document objects (e.g., the text, figures, tables, titles, and lists). The critical weakness of them is the low layout analysis quality for unseen journals and different document types. For example, when we apply the PubLayNet model trained with Detectron2 [1] to the first page of a Korean academic journal, it cannot capture the correct regions of documents objects, as depicted in Fig. 1.

In terms of training data and its coverage, PubLayNet and LayoutLM automatically construct the training data using the metadata provided by PubMed Central Open Access-eXtensible Markup Language (PMCOA-XML) or LaTex. Nevertheless, these are primarily for extracting figures and tables; they do not cover all the necessary metadata, such as the abstract, author name, keyword, or other data [1]. Moreover, to the best of our knowledge, the PMCOA-XML data of publications are only limited to biomedical journals and small numbers of LaTex data are available in the public domain. Recently, some training data for metadata extraction with consideration of the layout for the selected 40 Korean scientific journals were manually crafted [6]. However, its layout-aware data quality is not so satisfactory due to inconsistent and noisy annotations.

To guarantee consistent annotation quality in constructing layout-aware training data and building a more sophisticated language model for advanced metadata extraction, we propose a LAyout-aware MEtadata extraction (LAME) framework composed of three key components. First, an automatic layout analysis for metadata is designed with PDFMiner. Second, a large amount of layout-aware metadata is automatically constructed by analyzing the first page of papers in selected journals. Finally, Layout-aware

Bidirectional Encoder Representations from Transformers for Metadata (Layout-MetaBERT) models are constructed by adopting the BERT architecture [10].

In addition, to show the effectiveness of the Layout-MetaBERT models, we performed a set of experiments with other existing pretrained models and compared with the state-of-the-art (SOTA) model (i.e., bidirectional gated recurrent units and conditional random field (Bi-GRU-CRF)) for metadata extraction.

Our main contributions are as follows:

- We proposed an automatic layout analysis method which doesn't requires PMCOA-XML (or Latex) data for metadata extraction.
- We automatically generated training data for the layout-aware metadata from 70 research journals (65,007 PDF documents).
- We constructed a new pretrained language model, Layout-MetaBERT, to deal with the metadata of research articles more effectively.
- We demonstrated the effectiveness of Layout-MetaBERT on ten unseen research journals (13,300 PDF documents) with diverse layouts compared with the existing SOTA model (i.e., Bi-GRU-CRF).

## 2 Related work

### *2.1 Metadata extraction*

Various attempts have been made to analyze and extract information from documents and classify them into specific categories. Studies on text classification have been continuous since 1990, and the performance of text classification has gradually improved with the employment of sophisticated machine learning algorithms, such as the support vector machine (SVM) [11], conditional random fields (CRF) [12], convolutional neural network (CNN) [13], and bidirectional long short-term memory (BiLSTM) [14]. Afterward, various successful cases using bidirectional encoder representations from transformers (BERT) [10] pretrained with a large-scale corpus were introduced in the field of NLP. In the studies by [15] and [16], the pretrained BERT model was fine-tuned on the text classification task, and it showed results of close to or superior to the SOTA result for the target data. The use of BERT based pretrained models became popular due its high performances in various NLP fields, and more advanced pretrained models [17-19] were introduced according to various research purposes.

As a previous SOTA model of our metadata extraction task, a Bi-GRU-CRF model trained more than 20,000 human-annotated pages of layout boxes for metadata [6] from research articles showed an 82.46% of F1-score. However, accurately detecting and extracting regions for each type of metadata in documents is still a nontrivial task because of the various layout formats.

### *2.2 Document layout analysis*

Document layout analysis (DLA) [7] and several PDF handling efforts [6], [11], [20] have been conducted to understand the structure of documents. The DLA aims to identify the layout of text and nontext objects on the page and detect the layout function and format. Recently, the LayoutLM model [7] employed three different information elements for BERT pretraining to identify layouts: 1) layout coordinates, 2) text extracted using optical character recognition software, and 3) image embedding by understanding the layout structure through image processing. Moreover, NLP-based DLA research on various web documents [21], Layout detections and layout creation methods to find text information and location [8], [22], [23] have been studied. [2] and [24] applied the object detection technique to text region detection. Interestingly, widely used object detection techniques (e.g., Mask R-CNN [8] and Faster R-CNN [9]) have been applied to the metadata extraction field [1], [3].

Due to the high cost of training data construction for DLA, many studies have attempted to build datasets automatically. For example, the PubMed Central website, which includes academic documents in the biomedical field, provides a PMCOA-XML file for each document, enabling an analysis of the document structure. In the case of PubLayNet [1] which utilizes the PubMed dataset, the XML and PDFMiner's TextBoxes were matched to construct about 1 million training data. However, this is generally possible only when accurate coordinates are provided to separate each layout and the text information elements for each.

## 3 Proposed framework

Fig. 2 depicts our LAME framework consisting of three major components: automatic layout analysis, layout-aware training data construction, and Layout-MetaBERT generation.

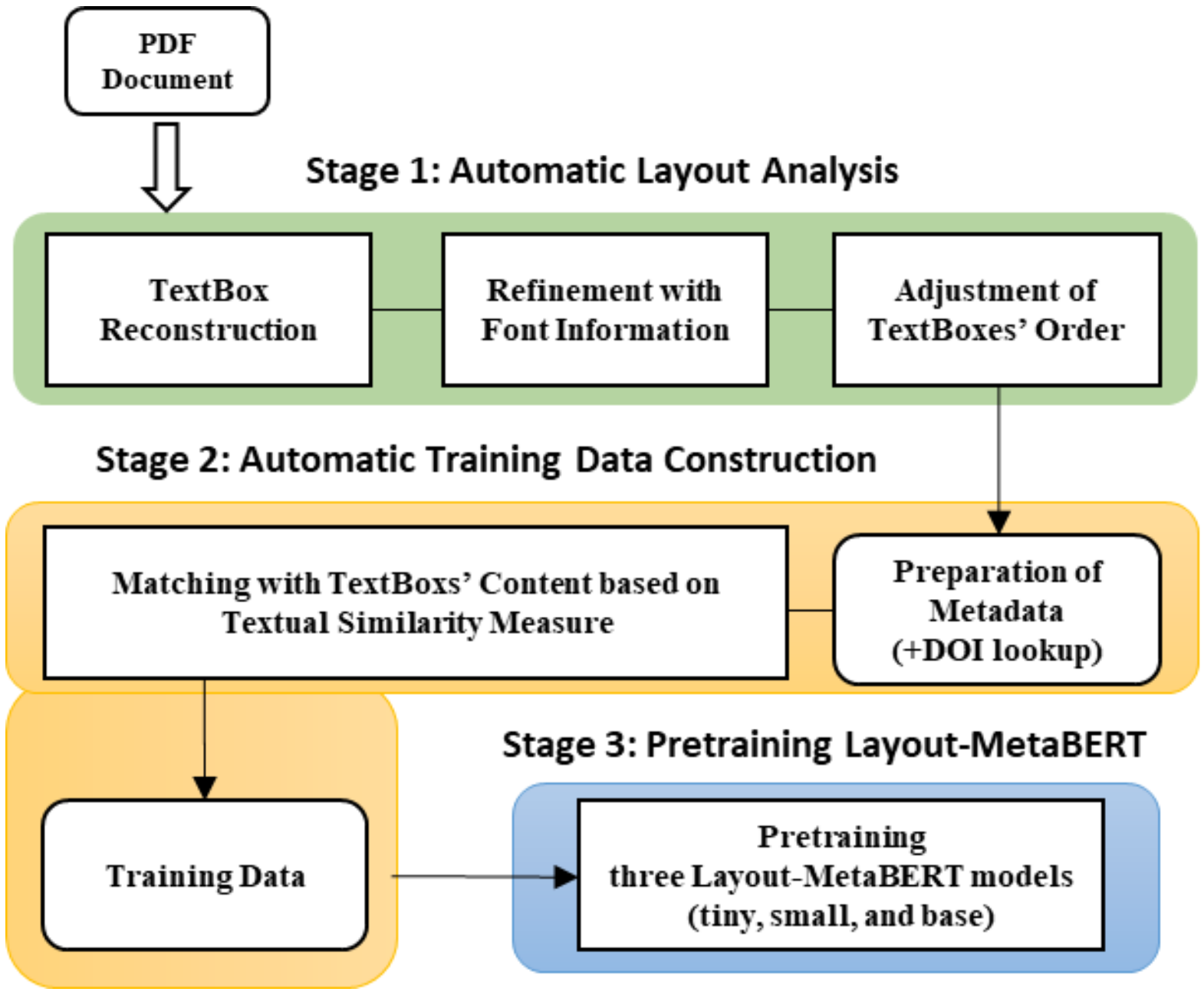

**Figure 2:** LAyout-Aware Metadata Extraction (LAME) framework.

### *3.1 Automatic layout analysis*

To understand the layout that separates each metadata element in the given PDF file, we must observe the text and coordinate information on the document's first page. To this ends, we employ the open-source software, PDFMiner, to extract meaningful information surrounding the text in the PDF files.

If we parse a PDF document with the software, we obtain information on the page, TextBox, TextLine, and character (Char) hierarchically, as illustrated in Fig. 3. These include various text information, such as coordinates, text, font size, and font for each object. For example, text coordinates appear in the form of (x, y) coordinates along with the height and width of the page.

#### *3.1.1 Textbox reconstruction*

To reduce existing errors in TextBox recognition, as depicted in Fig. 3, TextBoxes were reconstructed starting from the Char unit with the information obtained from PDFMiner. First, the spacing between characters is analyzed using the coordinate information for each Char. Generally, each token's x-coordinate distance (character spacing) appears the same, but the distance is slightly different depending on the alignment method or language.

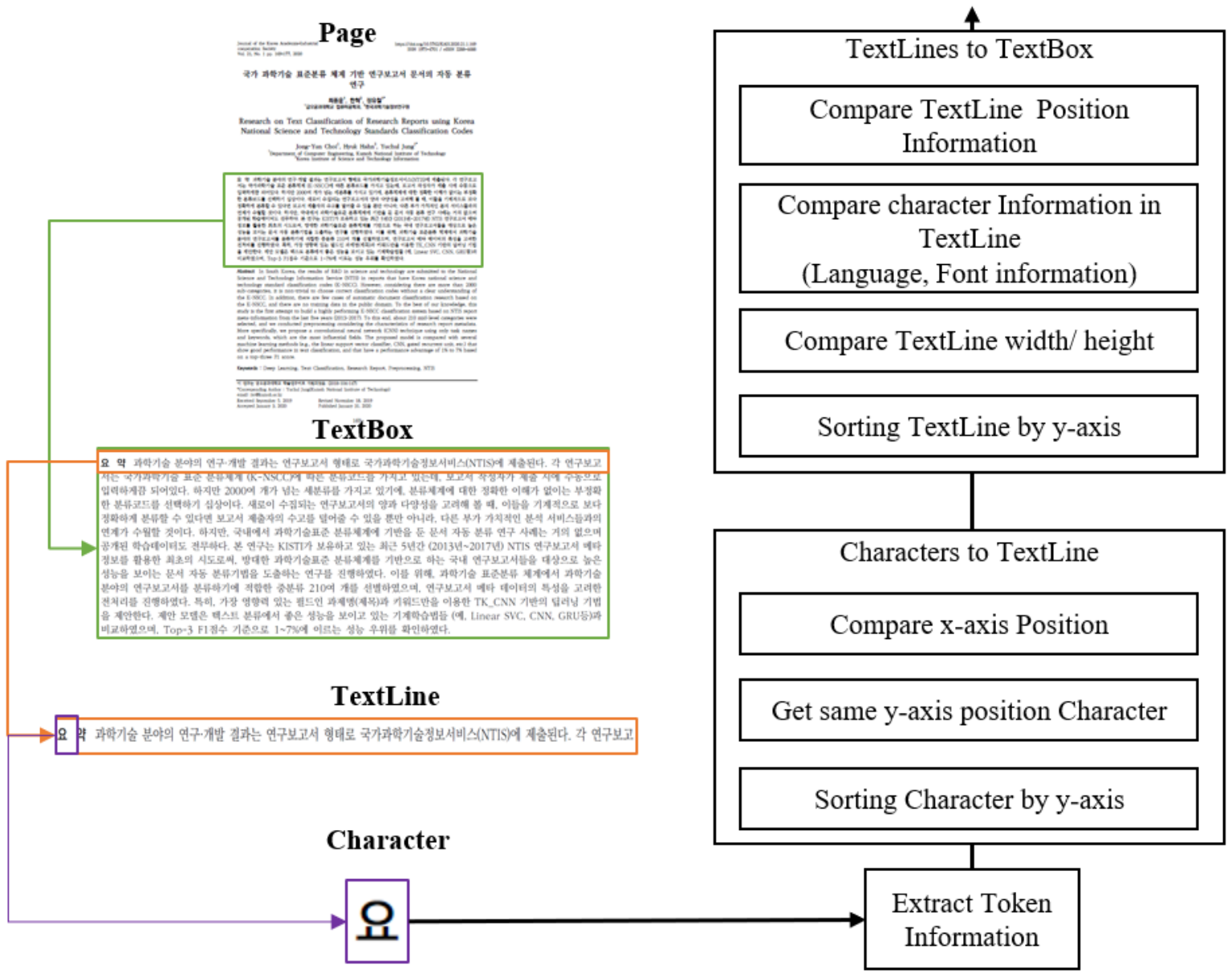

**Figure 3:** TextBox reconstruction based on the results of PDFMiner.

Therefore, after collecting characters in the same y-coordinate, the corresponding characters were sorted based on the x-coordinate value. As displayed in Fig. 4, the distance between each Char is smaller than the font size of the Chars; thus, the Char is determined to be part of the same TextLine in an academic document consisting of two columns. After aligning the TextLines based on the y-axis, if the distance between each y-coordinate is smaller than the height of each TextLine, the two different TextLines are regarded as the same TextBox.

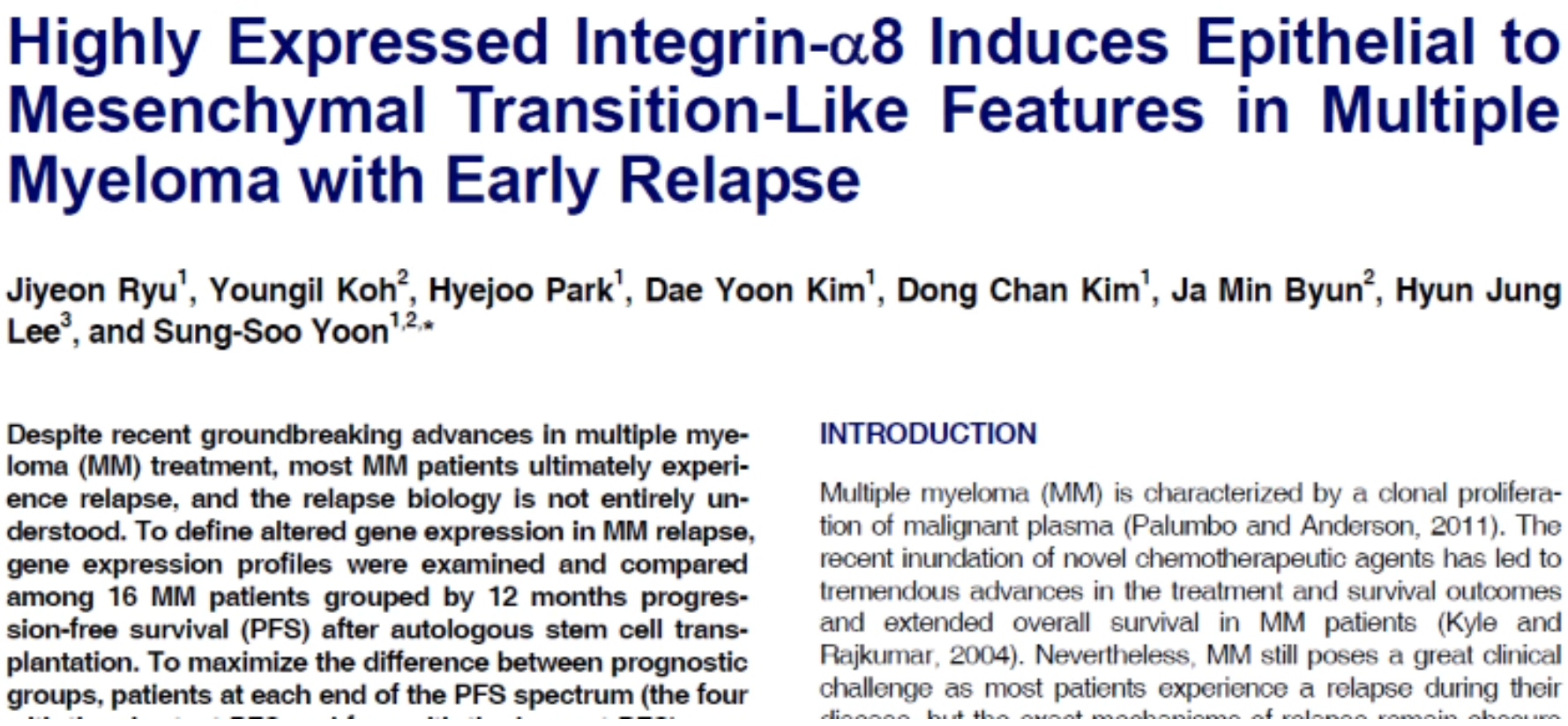
Highly Expressed Integrin-α8 Induces Epithelial to Mesenchymal Transition-Like Features in Multiple Myeloma with Early Relapse

Jiyeon Ryu[1], Youngil Koh[2], Hyejoo Park[1], Dae Yoon Kim[1], Dong Chan Kim[1], Ja Min Byun[2], Hyun Jung Lee[3], and Sung-Soo Yoon[1,2,*]

Despite recent groundbreaking advances in multiple myeloma (MM) treatment, most MM patients ultimately experience relapse, and the relapse biology is not entirely understood. To define altered gene expression in MM relapse, gene expression profiles were examined and compared among 16 MM patients grouped by 12 months progression-free survival (PFS) after autologous stem cell transplantation. To maximize the difference between prognostic groups, patients at each end of the PFS spectrum (the four

INTRODUCTION

Multiple myeloma (MM) is characterized by a clonal proliferation of malignant plasma (Palumbo and Anderson, 2011). The recent inundation of novel chemotherapeutic agents has led to tremendous advances in the treatment and survival outcomes and extended overall survival in MM patients (Kyle and Rajkumar, 2004). Nevertheless, MM still poses a great clinical challenge as most patients experience a relapse during their

**Figure 4:** Example of the separated column layout.

However, this method cannot create a TextBox accurately by separating paragraphs from paragraphs. For more elaborate TextBox composition, it needs to decide whether to configure the TextBox by considering the left x-coordinate $x_0$, the right x-coordinate $x_1$, and the width (W) of each TextLine. For example, for sentences like those in Fig. 5, we can think of two cases in composing a TextBox by comparing each TextLine.

First, the beginning of a paragraph is usually indented. Therefore, if the difference between the $x_0$ values of $L_i$ and $L_{i-1}$ is greater than the font size of the Chars existing in each TextLine, the two

TextLines should be included in different TextBoxes. Second, a TextLine that appears at the end of a paragraph has a shorter width because it has fewer Chars on average. Therefore, when the width of $L_{i-1}$ is smaller than the width of $L_i$, $L_{i-1}$ and $L_i$ should be assigned different TextBoxes.

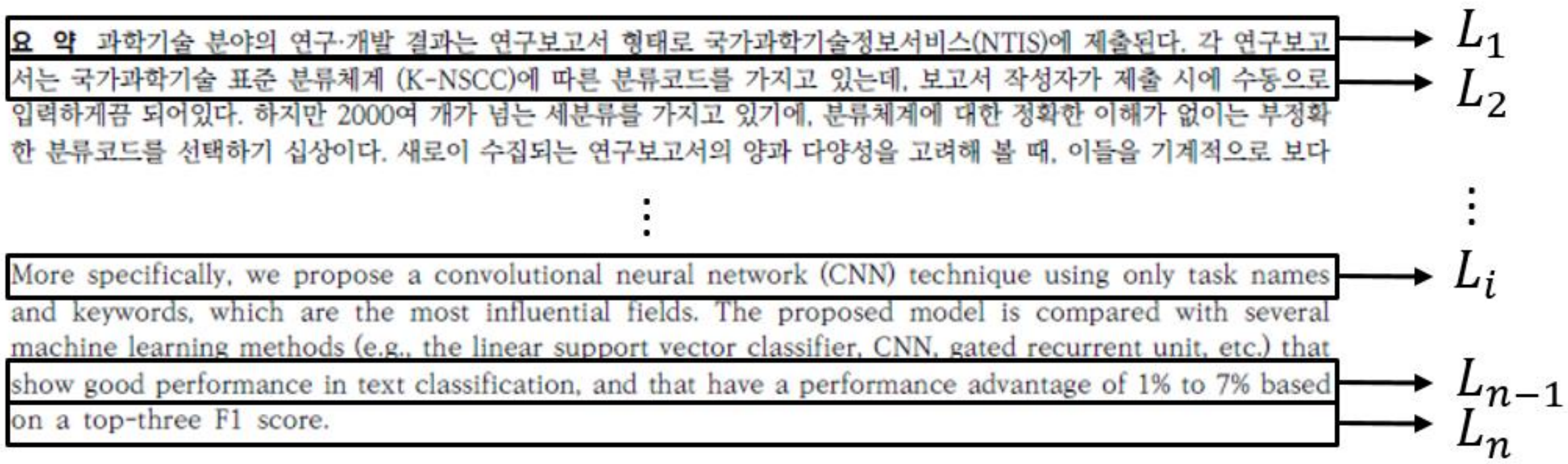

**Figure 5:** Example of TextBox separation.

### *3.1.2 Refinement with font information*

PDFMiner can produce various pieces of information in terms of font information, such as the font name and style (e.g., bold, italic, etc.) as listed in Table 1.

**Table 1:** Example of font information when pdfminer is applied

| Text | PDFMiner output font name |
|---|---|
| Abstract | ELNFKM+KoreanGD-Bold-KSCpc-EUC-H |
| Abstract | INPILL+Gulim |
| ***Abstract*** | Arial-BoldItalicMT |
| ABSTRACT | GPJCIE+YDIYGO130 |

However, as in Fig. 6, English is frequently used for Korean abstracts in some journals published in Korea. In particular, abstracts written in Korean and English appear together on the first page of some research articles. In addition, certain strings are often treated as bold or italic and often have different fonts and sizes, such as section titles. Considering this problem, when composing a TextBox using the coordinate information described above, if the font information displayed on each line is different, it is not simply judged as a different line. After analyzing the font information of different languages that appear, the TextBox was determined by considering the number of the appearing fonts (e.g., bold and italic).

Although font information helps the layout composition, it is still confusing when the same font information is used for individual information marking or bold processing for emphasis or different metadata. Additional processing is required to correctly connect individual fonts to make a layout using the font information. Therefore, we compared only texts described in Korean and English and used only the fonts of the same language to determine the layout.

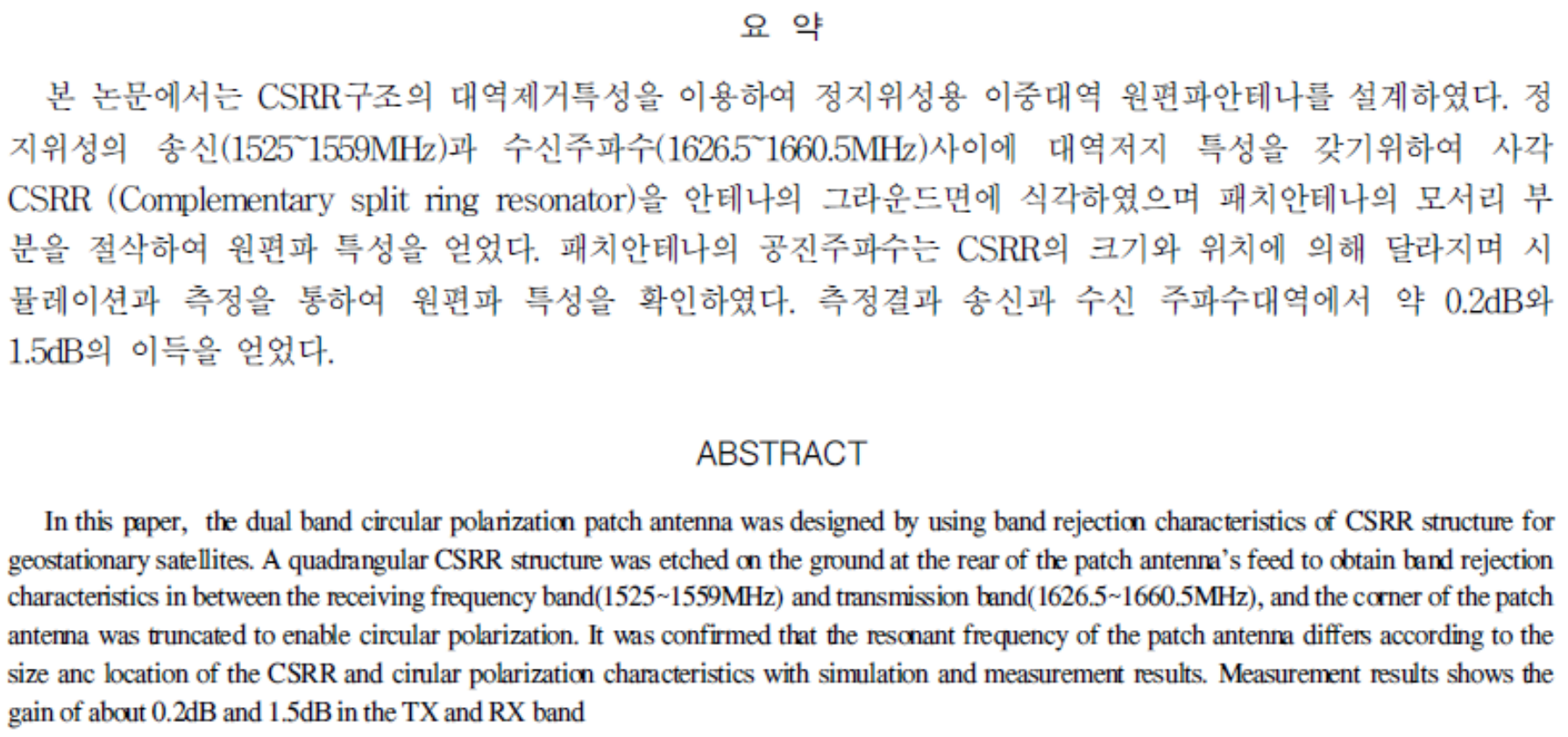
요 약

본 논문에서는 CSRR구조의 대역제거특성을 이용하여 정지위성용 이중대역 원편파안테나를 설계하였다. 정지위성의 송신(1525~1559MHz)과 수신주파수(1626.5~1660.5MHz)사이에 대역저지 특성을 갖기위하여 사각 CSRR (Complementary split ring resonator)을 안테나의 그라운드면에 식각하였으며 패치안테나의 모서리 부분을 절삭하여 원편파 특성을 얻었다. 패치안테나의 공진주파수는 CSRR의 크기와 위치에 의해 달라지며 시뮬레이션과 측정을 통하여 원편파 특성을 확인하였다. 측정결과 송신과 수신 주파수대역에서 약 0.2dB와 1.5dB의 이득을 얻었다.

ABSTRACT

In this paper, the dual band circular polarization patch antenna was designed by using band rejection characteristics of CSRR structure for geostationary satellites. A quadrangular CSRR structure was etched on the ground at the rear of the patch antenna's feed to obtain band rejection characteristics in between the receiving frequency band(1525~1559MHz) and transmission band(1626.5~1660.5MHz), and the corner of the patch antenna was truncated to enable circular polarization. It was confirmed that the resonant frequency of the patch antenna differs according to the size anc location of the CSRR and cirular polarization characteristics with simulation and measurement results. Measurement results shows the gain of about 0.2dB and 1.5dB in the TX and RX band

**Figure 6:** Example of when a Korean abstract and an English abstract exist together.

#### *3.1.3 Adjustment of text box order*

Academic papers may consist of one or two columns depending on the format for each journal. In some cases, only the main body consists of two columns, and the title, abstract, and author name are displayed in one column. For example, in Fig. 3, such information as the title and author name was arranged in the center, but the document object identifier (DOI) information or academic journal names appeared separately on the left and right sides. To effectively identify metadata consistently from varied layout formats, we sorted the textboxes extracted from the first pages of the research articles sequentially from top to bottom based on the y-axis.

### *3.2 Automatic training data construction*

We compared the content extracted from the PDFMiner with the metadata prepared in advance to construct the layout-aware metadata automatically. If no metadata is available for the given research article, metadata can be automatically obtained through the DOI lookup. Therefore, this technique can be extended to all journal types where the registered DOI exists.

However, the compared textual content is not always precisely matched. Therefore, to determine the extent of the match, we allowed only fields with almost identical (or high similarity) matches for each layout text information element automatically acquired in the previous step as training data. We used a mixed textual-similarity measure for efficient computation based on the Levenshtein distance and bilingual evaluation understudy (BLEU) score.

The Levenshtein distance was calculated using Python's fuzzywuzzy[1]. The scores calculated using the BLEU [25] measure were summed to determine whether the given metadata displays a degree of agreement of 80% or more. Nevertheless, some post-processing is required in the process. In analyzing the text after extraction, some problems occur when dealing with expression substitutions (e.g., "<TEX>," cid:0000). Encoding errors reduced the portion of mathematical expressions that can be removed as much as possible, and we excluded the text with encoding problems to avoid these errors.

### *3.3 Pretraining Layout-MetaBERT*

Although pretraining a BERT model requires a large corpus and a long training time, a fine-tuning step can make a difference in performance depending on the characteristics of the data used for pretraining. For example, when pretrained with specific domain data, such as SciBERT [19] and BioBERT [26], they performed better than Google's BERT model [10] in downstream tasks of science and technology or medical fields. However, to our best knowledge, there is no pretrained model designed to extract metadata based on research article data.

---

[1] https://github.com/seatgeek/fuzzywuzzy

Therefore, we newly pretrained a layout-aware language model, so called, Layout-MetaBERT, that can effectively deal with metadata from research articles. Fig. 7 describes how the previously constructed training data are used for pretraining and fine-tuning the Layout-MetaBERT.

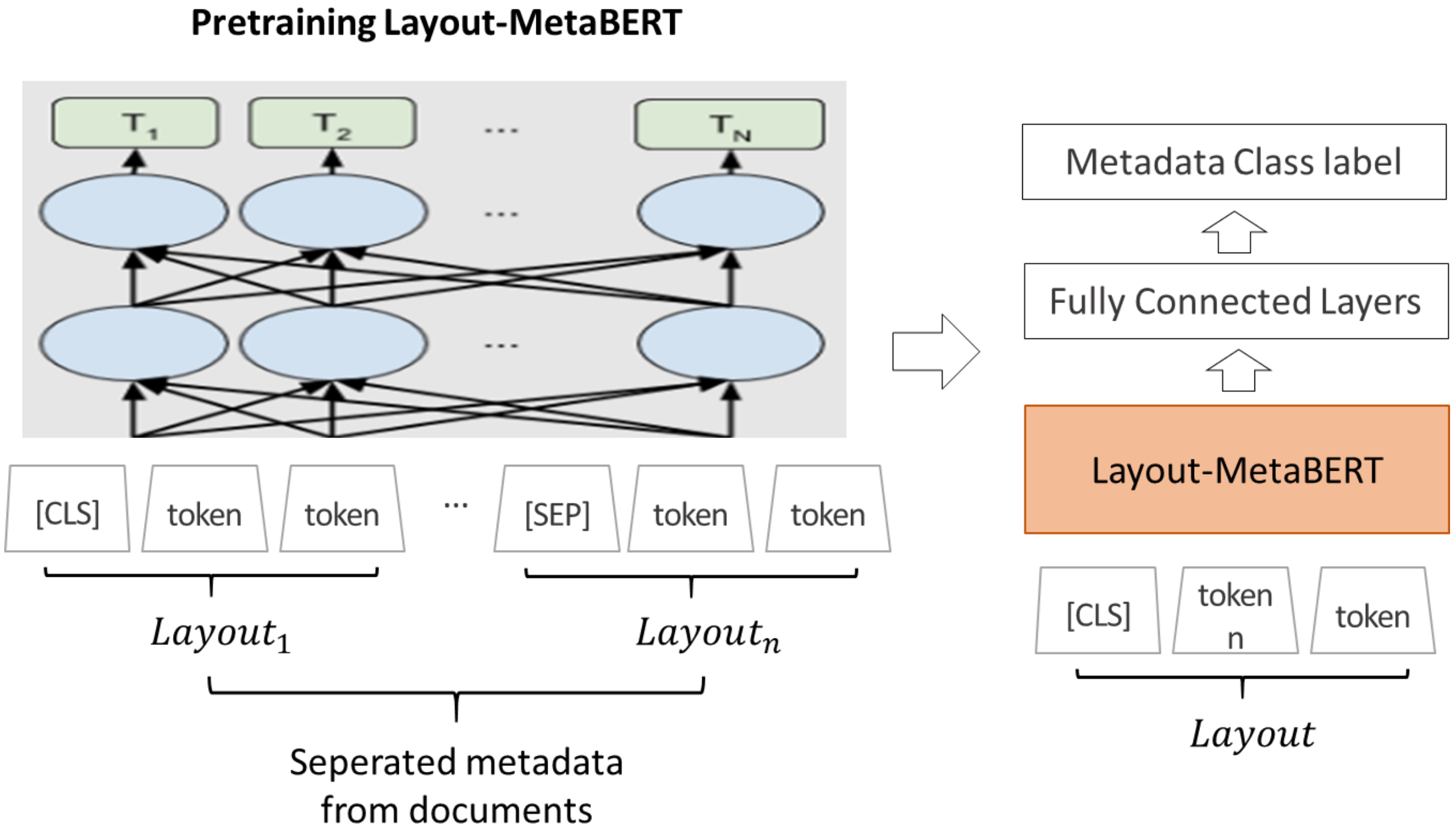

**Figure 7:** Pretraining Layout-MetaBERT and fine-tuning for classification downstream task.

Different from the Google BERT model [10], in our Layout-MetaBERT pretraining, each document layout was considered a sequence in this study. Thus, each layout was classified by the [SEP] token to prepare the training data and was used for pretraining. In pretraining Layout-MetaBERT models, we followed three size models of the Google BERT: base (L = 12, H = 768, A = 12), small (L = 4, H = 512, A = 8), and tiny (L = 2, H = 128, A = 2), where L is the number transformer blocks, H is the hidden size, and A is the number of self-attention heads. We used a dictionary of 10,000 words built through the WordPiece mechanism and automatically generated training data extracted from the first page of 60 research journals among the 70 journals for the pretraining. The pretrained Layout-MetaBERT can be used for metadata extraction after fine-tuning.

## 4 Experiments

We summarize the results of three major components to examine the applicability of the LAME framework. First, we compare the results of the proposed automatic layout analysis with other layout analysis techniques. Second, we describe the statistics of the training data constructed according to the results of the automatic layout analysis. Finally, we compare metadata extraction performances our constructed Layout-MetaBERT models with other deep learning and machine learning techniques after fine-tuning.

### *4.1 Comparison with other layout analysis methods*

No prepared correct answers exist for the target research articles; thus, we compared the generated layout boxes from PDFMiner, PubLayNet, and the proposed layout analysis method for two randomly selected documents (e.g., A and B) as depicted in Fig. 8. In Document A, the extraction results of PDFMiner and the layout analysis are similar. However, in PubLayNet, the information of the paragraphs is excessively separated, as indicated in Fig. 8-(a). For Document B, the extraction results of all techniques were somewhat similar, but PubLayNet displayed the author name and affiliation as one piece of information, and PDFMiner produced a separate box for the line-wrapped title.

The proposed method could generate a good enough layout analysis for the first page of the research articles through the comparisons. Comparing all three layout analysis results manually for each layout box to calculate the accuracy requires too much human labor and is beyond the scope of this paper. The performance of the constructed Layout-MetaBERT indirectly measures the quality of the layout analysis.

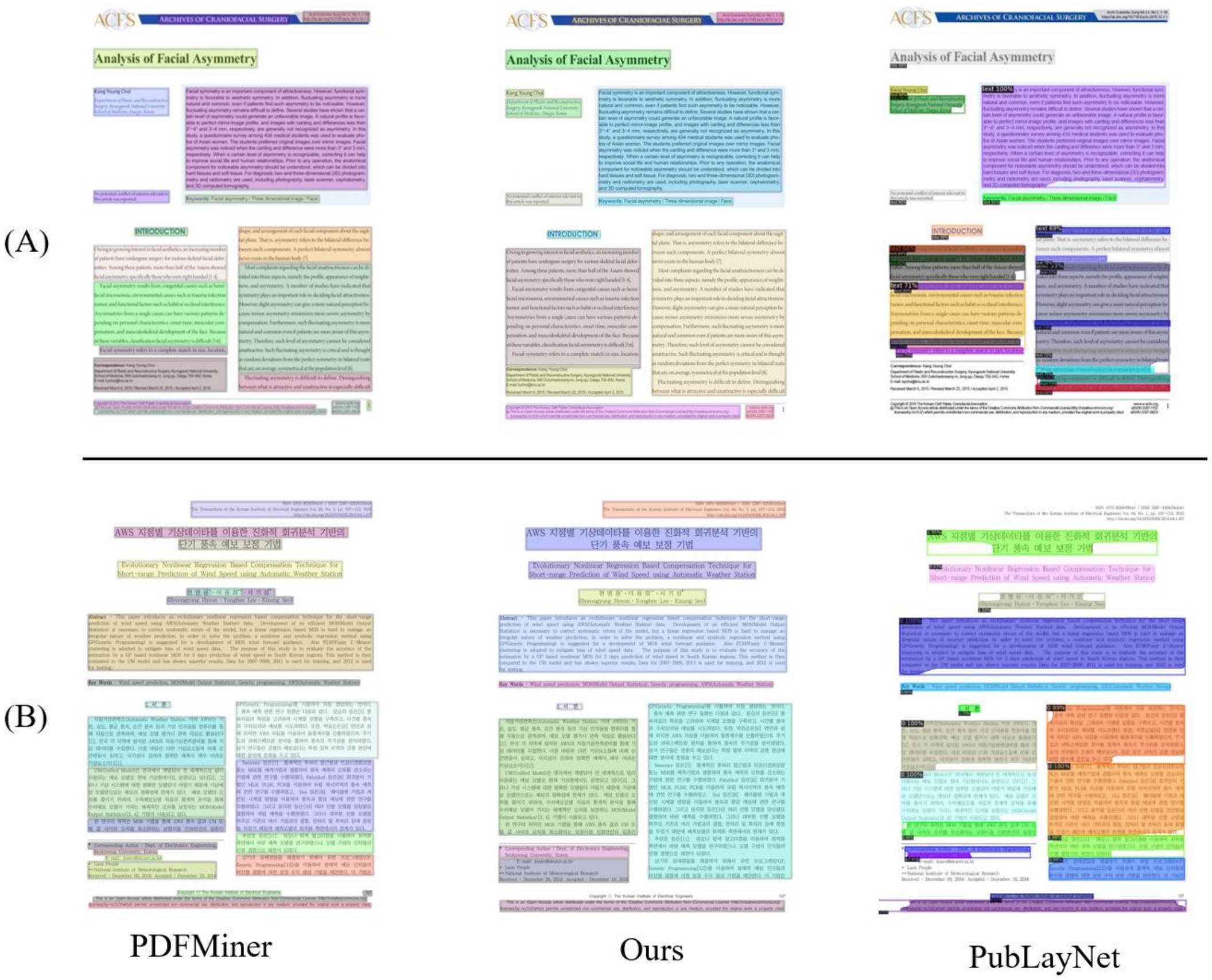

**Figure 8:** Layout analysis comparisons with PDFMiner and PubLayNet.

### *4.2 Training data construction*

To reflect various kinds of layout formats, we used 70 research journals (Appendix 1) provided by the Korea Institute of Science and Technology Information (KISTI) to extract major metadata elements, such as titles, author names, author affiliations, keywords, and abstracts in Korean and English based on the automatic layout analysis in Section 3.1. Among the 70 journals, two journals were written in only Korean, 23 journals in only English, and 45 in Korean and English.

For each layout that separates metadata on the first page of the 70 journals (65,007 PDF documents), automatic labeling with ten labels was performed, and other layouts not included in the relevant information were labeled O. The statistics of automatically generated training data are presented in Table 2.

**Table 2** Statistics for automatically generated training data

| Metadata Field | Label (i.e., layout) | Count |
|---|---|---|
| Out of boundary | O | 637,856 |
| Title (in Korean) | title_ko | 46,056 |
| Title (in English) | title_en | 64,414 |
| Affiliation (in Korean) | org_ko | 39,233 |
| Affiliation (in English) | org_en | 63,434 |
| Abstract (in Korean) | abstract_ko | 31,885 |
| Abstract (IN English) | abstract_en | 55,318 |
| Keywords (in Korean) | keywords_ko | 21,685 |
| Keywords (in English) | keywords_en | 61,221 |
| Author name (in Korean) | author_name_ko | 56,306 |
| Author name (in English) | author_name_en | 35,631 |

***4.3 Experimental results***

To check the performance of the proposed Layout-MetaBERT, 70 research journals (65,007 documents) were divided into 60 (51,676 documents) for pretraining (and fine-tuning) and 10 (13,331 documents) for testing, respectively. Table 3 lists the training and testing performances of the three Layout-MetaBERT models with widely used metadata extraction techniques. Finally, Table 4 describes the Macro-F1 and Micro-F1 scores for metadata classification comparisons with existing pretrained models.

***4.3.1 Fine-tuning and Hyperparameters***

In fine-tuning with various pretrained language models (e.g., three different sized models of Layout-MetaBERT, KoALBERT, KoELECTRA, and KoBERT), all experiments were conducted under the same configurations with an epoch of 5, batch size of 32, learning rate of 2e-5, and maximum sequence length of 256. In addition, we used the Nvidia RTX Titan 4-way system and Google's TensorFlow framework in Python 3.6.9 for pretraining and fine-tuning.

***4.3.2 Stable performances of Layout-MetaBERT***

The proposed Layout-MetaBERT models can effectively extract metadata, as listed in Table 3. In particular, Layout-MetaBERT models make significant differences compared to the existing SOTA (i.e., Bi-GRU-CRF) model. Even the tiny model with the fewest parameters among the Layout-MetaBERT models has higher performance than other pretrained models in Macro-F1 and Micro-F1 scores, as displayed in Table 4. Moreover, three Layout-MetaBERT models have only minor differences between the Micro-F1 and Macro-F1 scores compared to other pretrained models. Moreover, the Layout-MetaBERT models exhibit 90% or more robustness in metadata extraction, confirming that pretraining the layout units with the BERT schemes is feasible in the metadata extraction task.

**Table 3:** Train and test performances of metadata extraction

| Models | Micro-F1 (train) | Micro-F1 (test) |
|---|---|---|
| Layout-MetaBERT (base) | 0.9559 | 0.936 |
| Layout-MetaBERT (small) | 0.9595 | 0.9333 |
| Layout-MetaBERT (tiny) | 0.9432 | 0.9293 |
| KoBERT[2] | 0.8086 | 0.7901 |
| KoalBERT[3] | 0.9014 | 0.8978 |
| KoELECTRA[4] | 0.9354 | 0.9204 |

[2] https://github.com/SKTBrain/KoBERT

[3] https://huggingface.co/kykim/albert-kor-base

| | | |
|---|---|---|
| Bi-GRU-CRF [6] (without position) | 0.8610 | 0.8912 |
| Bi-GRU-CRF [6] (with position) | 0.9442 | 0.0985 |
| CNN [13] | 0.9425 | 0.824 |
| SVM [11] | 0.9411 | 0.8114 |

**Table 4:** Metadata extraction performances of primary BERT models for each label

| | Layout-MetaBERT (tiny) | Layout-MetaBERT (small) | Layout-MetaBERT (base) | KoBERT | KoALBERT | KoELECTRA |
|---|---|---|---|---|---|---|
| Model size | 5M | 16M | 110M | 110M | 12M | 110M |
| O | 0.94 | 0.94 | 0.95 | 0.85 | 0.92 | 0.94 |
| title_ko | 0.92 | 0.94 | 0.93 | 0.59 | 0.89 | 0.91 |
| title_en | 0.92 | 0.91 | 0.92 | 0.76 | 0.8 | 0.87 |
| org_ko | 0.96 | 0.96 | 0.96 | 0.36 | 0.96 | 0.94 |
| org_en | 0.86 | 0.87 | 0.9 | 0.64 | 0.79 | 0.9 |
| abstract_ko | 0.92 | 0.93 | 0.93 | 0.84 | 0.9 | 0.92 |
| abstract_en | 0.92 | 0.93 | 0.94 | 0.84 | 0.91 | 0.91 |
| keywords_ko | 0.94 | 0.95 | 0.95 | 0.86 | 0.94 | 0.95 |
| keywords_en | 0.87 | 0.89 | 0.9 | 0.51 | 0.91 | 0.73 |
| author_name_ko | 0.97 | 0.97 | 0.96 | 0.75 | 0.95 | 0.92 |
| author_name_en | 0.56 | 0.62 | 0.92 | 0.3 | 0.41 | 0.57 |
| Micro-F1 | 0.9293 | 0.9333 | 0.9360 | 0.7901 | 0.8978 | 0.9204 |
| Macro-F1 | 0.8891 | 0.9009 | 0.9327 | 0.6636 | 0.8527 | 0.8691 |

#### *4.3.3 Experiments with position information*

Unlike other models, the Bi-GRU-CRF model used the absolute coordinates of metadata with other textual features. However, the model failed to discriminate unseen layouts from unseen journals when using the coordinate information for training various journal layout formats. Therefore, to determine the validity of the coordinate information, we performed additional experiments with the Bi-GRU-CRF (with position) and Bi-GRU-CRF (without position) models. Although Bi-GRU-CRF (with position) model demonstrated high performance in the training stage, it failed to recognize metadata-related layouts in unseen journals (less than 10% as F1 score). However, the performance of the Bi-GRU-CRF (without position) model had somewhat lower performance in the training stage compared to the other models. The model performed well, similar with that of KoALBERT. Thus, we confirmed that using absolute coordinate information can only be applied under the premise that the journals used in training also are used in testing.

## 5 Discussion

### *5.1 Additional performance improvements*

The proposed Layout-MetaBERT exhibited higher results than the existing SOTA model [6]. However, absolute coordinate information could obtain poor results for documents in a format not learned. In addition, the proposed layout analysis method separates the metadata well from the first page of the academic documents of various layouts.

[4] https://github.com/monologg/KoELECTRA

However, the accuracy of the automatically generated training data is not perfect. There may be errors due to the difference between the metadata format of the document and the metadata written in advance. As mentioned, encoding errors also occur in extracting text from mathematical formulas or PDF documents. Generating the correct layout has a significant effect on extracting metadata and is an essential factor in automatically generating data. Therefore, if more sophisticated training data can be generated, the performance of Layout-MetaBERT can be further improved.

### *5.2 Restrictions of Layout-MetaBERT*

Much research has been conducted on automatically extracting layouts from PDF documents. Creating accurate layouts has a significant influence on meta-extraction. This study attempted to compose the layout of the first page of an academic document using text information. Based on this, we trained the Layout-MetaBERT and confirmed the positive results for the applicability to the meta classification module. However, the proposed technique cannot be applied to all documents. An image-type PDF cannot be used unless the text is extracted. In this case, the extraction must be performed using a high-performance optical character recognition module.

### *5.3 Expansion to other metadata types*

This study focused on extracting five major metadata elements (i.e., titles, abstracts, keywords, author names, and author information). Considering that the target research articles contain elements written in English, Korean, or both, the number of metadata becomes 10. However, other metadata (e.g., publication year, start page, end page, DOI, volume number, journal title, etc.) can be extracted further by applying highly refined regular expressions in the post-processing step.

## 6 Conclusion

In this paper, the LAME framework is proposed to extract metadata from PDFs of research articles with high performance. First, the automatic layout analysis detects the layout regions where metadata exists regardless of the journal formats based on text features, text coordinates, and font information. Second, by constructing automatic training data, we built high-quality metadata-separated training data for 70 journals (65,007 documents). In addition, our fine-tuned Layout-MetaBERT (base) demonstrated excellent metadata extraction performance (F1 = 94.6%) for even unseen journals with diverse layouts. Moreover, Layout-MetaBERT (tiny) with the fewest parameters exhibited superior performance than other pretraining models, implying that well-separated layouts induce effective metadata extraction when they meet appropriate language models.

In future work, we plan to conduct experiments to determine whether the proposed model applies to the more than 500 other journals not used in this study. Moreover, resolving potential errors in the automatically generated training data is a concern to create layouts that separate each metadata element in an advanced way. Furthermore, extending the number of metadata items extracted without post-processing is an exciting but challenging task to resolve as future work.

**Conflicts of Interest:** The authors declare that they have no conflicts of interest to report regarding the present study.

## References

[1] X. Zhong, J. Tang and A. J. Yepes, "Publaynet: largest dataset ever for document layout analysis," in *International Conference on Document Analysis and Recognition (ICDAR). IEEE*, 2019.

[2] L. Melinda, R. Ghanapuram, and C. Bhagvati, "Document layout analysis using multigaussian fitting," in *14th IAPR International Conference on Document Analysis and Recognition (ICDAR)*, vol. 1, pp. 747-752, 2017.

[3] M. Li et al., "DocBank: A Benchmark Dataset for Document Layout Analysis," in *28th International Conference on Computational Linguistics*, pp. 949-960, 2020.

[4] D. Tkaczyk, P. Szostek, M. Fedoryszak and P. Jan, "CERMINE: automatic extraction of structured metadata from scientific literature," in *International Journal on Document Analysis and Recognition (IJDAR),* vol. 18, no. 4, pp. 317-335, 2015.

[5] P. Lopez, "GROBID: Combining automatic bibliographic data recognition and term extraction for scholarship publications," in *International conference on theory and practice of digital librarie*, Springer, Berlin, Heidelberg, 2009.

[6] S. Kim, S. Ji, H. Jeong, H. Yoon and S. Choi, "Metadata Extraction based on Deep Learning from Academic Paper in PDF," *Journal of KIISE,* vol. 46, no. 7, pp. 644–652, 2019.

[7] Y. Xu, M. Li, L. Cui, S. Huang, F. Wei and M. Zhou, "Layoutlm: Pre-training of text and layout for document image understanding," in *Proc. 26th ACM SIGKDD International Conference on Knowledge Discovery & Data Mining*, pp. 1192-1200, 2020.

[8] K. He, G. Gkioxari, P. Dollár and R. Girshick, "Mask r-cnn," in *Proc. IEEE international conference on computer vision*, pp. 2961-2969, 2017.

[9] R. Girshick, "Fast r-cnn," in *Proc. IEEE international conference on computer vision*, pp. 1440-1448, 2015.

[10] J. Devlin, M. W. Chang, K. Lee and K. Toutanova, "BERT: Pre-training of Deep Bidirectional Transformers for Language Understanding," in *Proc. 2019 Conference of the North American Chapter of the Association for Computational Linguistics: Human Language Technologies*, vol. 1, pp. 4171-4186, 2019.

[11] H. Han, C.L. Giles, E. Manavoglu, H. Zha and E.A. Fox, "Automatic document metadata extraction using support vector machines," *Joint Conference on Digital Libraries*, pp. 37-48, 2003.

[12] M. Abramson, "Sequence classification with neural conditional random fields," in *IEEE 14th International Conference on Machine Learning and Applications (ICMLA)*, pp. 799-804, 2015.

[13] Y. Kim, "Convolutional Neural Networks for Sentence Classification," MS thesis, University of Waterloo, 2015.

[14] P. Zhou, Z. Qi, S. Zheng, J. Xu and H. Bao et al., "Text Classification Improved by Integrating Bidirectional LSTM with Two-dimensional Max Pooling," in *Proc. COLING 2016, the 26th International Conference on Computational Linguistics: Technical Papers,* pp. 3485-3495, 2016.

[15] A. Adhikari, A. Ram, R. Tang, and J. Lin, ''DocBERT: BERT for document classification,'' arXiv preprint arXiv:1904.08398, 2019, [Online]. Available: http://arxiv.org/abs/1904.08398

[16] S. Yu, J. Su and D. Luo, "Improving bert-based text classification with auxiliary sentence and domain knowledge," *IEEE Access*, vol. 7, pp. 176600-176612, 2019.

[17] X. Gu, K. M. Yoo, and J. Ha, "DialogBERT: Discourse-Aware Response Generation via Learning to Recover and Rank Utterances," arXiv preprint arXiv:2012.01775v1, 2021, [Online]. Available: https://arxiv.org/abs/2012.01775

[18] Z. Lan, M. Chen, S. Goodman, K. Gimpel and P. Sharma et al., "ALBERT : A L ITE BERT FOR S ELF - SUPERVISED," *International Conference on Learning Representations*, 2019.

[19] I. Beltagy, K. Lo, and A. Cohan, "SciBERT: A Pretrained Language Model for Scientific Text," in *Proc 2019 Conference on Empirical Methods in Natural Language Processing and the 9th International Joint Conference on Natural Language Processing (EMNLP-IJCNLP)*, pp. 3615-3620, 2019.

[20] A. R. Katti, C. Reisswig, C. Guder, S. Brarda and S. Bickel et al., "Chargrid: Towards Understanding 2D Documents," *in Proc 2018 Conference on Empirical Methods in Natural Language Processing*, pp.4459-4469, 2018.

[21] Ł. Garncarek, R. Powalski, and T. Stanisławek, "LAMBERT: Layout-Aware (Language) Modeling for information extraction," in arXiv preprint arXiv:2002.08087, 2020, [Online]. Available: https://arxiv.org/abs/2002.08087

[22] Z.-Q. Zhao, P. Zheng, S. Xu and X. Wu, "Object detection with deep learning: A review," *IEEE transactions on neural networks and learning systems*, vol. 30, no. 11, pp. 3212-3232, 2019.

[23] Y. Liu, L. Jin, Z. Xie, C. Luo and S. Zhang et al., "Tightness-aware evaluation protocol for scene text detection," in *Proc IEEE/CVF Conference on Computer Vision and Pattern Recognition*, pp. 9612-9620, 2019.

[24] A. Simon, J.-C. Pret and A.P. Johnson, "A fast algorithm for bottom-up document layout analysis," *IEEE Transactions on Pattern Analysis and Machine Intelligence*, vol. 19, no. 3, pp. 273-277, 1997.

[25] K. Papineni, S. Roukos, T. Ward, and W. J. Zhu, "Bleu: a method for automatic evaluation of machine translation," in *Proc 40th annual meeting of the Association for Computational Linguistics*, pp. 311-318, 2002.

[26] J. Lee, W. Yoon, S. Kim, D. Kim, and S. Kim, et al., "BioBERT: a pre-trained biomedical language representation model for biomedical text mining," *Bioinformatics*, vol. 36, no. 4, pp. 1234-1240, 2020.

**<Appendix A>**

| JOURNALS | NUMBERS OF SELECTED PAPERS |
|---|---|
| TRAIN SET | |
| JOURNAL OF THE KOREAN CLEFT PALATE-CRANIOFACIAL ASSOCIATION | 248 |
| KOREAN JOURNAL OF CONSTRUCTION ENGINEERING AND MANAGEMENT | 575 |
| JOURNAL OF INTERNET COMPUTING AND SERVICES | 513 |
| JOURNAL OF THE KOREAN SOCIETY OF RADIOLOGY | 726 |
| JOURNAL OF THE KOREA INSTITUTE OF INFORMATION AND COMMUNICATION ENGINEERING | 2477 |
| KOREAN JOURNAL OF MATERIALS RESEARCH | 892 |
| FOOD SCIENCE OF ANIMAL RESOURCES | 636 |
| KOREAN JOURNAL OF PEDIATRICS | 812 |
| KOREAN CHEMICAL ENGINEERING RESEARCH | 761 |
| JOURNAL OF INFORMATION PROCESSING SYSTEMS | 97 |
| JOURNAL OF DIGITAL CONVERGENCE | 3351 |
| JOURNAL OF THE KOREA CONVERGENCE SOCIETY | 1411 |
| JOURNAL OF THE KOREAN SOCIETY OF CLOTHING AND TEXTILES | 621 |
| MOLECULES AND CELLS | 360 |
| JOURNAL OF KOREAN ACADEMY OF NURSING | 772 |
| JOURNAL OF THE KOREAN SOCIETY OF INTEGRATIVE MEDICINE | 310 |
| THE JOURNAL OF THE INSTITUTE OF INTERNET, BROADCASTING AND COMMUNICATION | 1558 |
| JOURNAL OF KOREA WATER RESOURCES ASSOCIATION | 727 |
| BULLETIN OF THE KMS | 1133 |
| INTERNATIONAL JOURNAL OF ADVANCED SMART CONVERGENCE | 341 |
| ARCHIVES OF PLASTIC SURGERY | 1135 |
| THE JOURNAL OF THE KOREA INSTITUTE OF ELECTRONIC COMMUNICATION SCIENCES | 852 |
| APPLIED CHEMISTRY FOR ENGINEERING | 537 |
| JOURNAL OF KOREA INSTITUTE OF INFORMATION, ELECTRONICS, AND COMMUNICATION TECHNOLOGY | 448 |
| PEDIATRIC GASTROENTEROLOGY, HEPATOLOGY & NUTRITION | 373 |
| THE JOURNAL OF THE KOREA CONTENTS ASSOCIATION | 5524 |
| THE JOURNAL OF KOREAN ORIENTAL INTERNAL MEDICINE | 603 |
| KSII TRANSACTIONS ON INTERNET AND INFORMATION SYSTEMS (TIIS) | 1256 |
| JOURNAL OF KOREAN MEDICINE REHABILITATION | 250 |
| THE KOREAN JOURNAL OF PHYSIOLOGY AND PHARMACOLOGY | 518 |
| THE JOURNAL OF KOREAN PHYSICAL THERAPY | 561 |
| JOURNAL OF PREVENTIVE MEDICINE AND PUBLIC HEALTH | 420 |
| THE KOREAN JOURNAL OF THORACIC AND CARDIOVASCULAR SURGERY | 779 |
| CHILD HEALTH NURSING RESEARCH | 393 |
| BIOMOLECULES & THERAPEUTICS | 580 |
| ASIAN-AUSTRALASIAN JOURNAL OF ANIMAL SCIENCES | 417 |
| JOURNAL OF ENVIRONMENTAL HEALTH SCIENCES | 533 |
| THE KOREAN JOURNAL OF PARASITOLOGY | 705 |
| JOURNAL OF THE KOREAN SOCIETY OF PHYSICAL MEDICINE | 547 |
| JOURNAL OF THE KOREA INSTITUTE OF INFORMATION SECURITY AND CRYPTOLOGY | 919 |
| JOURNAL OF THE KOREAN ASSOCIATION FOR SCIENCE EDUCATION | 638 |
| JOURNAL OF THE KOREAN APPLIED SCIENCE AND TECHNOLOGY | 747 |
| JOURNAL OF THE KOREAN SOCIETY OF CIVIL ENGINEERS | 1226 |
| JOURNAL OF DIGITAL CONTENTS SOCIETY | 580 |
| THE KOREAN JOURNAL OF FOOD AND NUTRITION | 967 |
| JOURNAL OF KOREAN NEUROSURGICAL SOCIETY | 1132 |
| JOURNAL OF MICROBIOLOGY AND BIOTECHNOLOGY | 1281 |
| THE KOREAN JOURNAL OF APPLIED STATISTICS | 591 |
| JOURNAL OF POWER ELECTRONICS | 154 |
| MICROBIOLOGY AND BIOTECHNOLOGY LETTERS | 463 |
| THE JOURNAL OF KOREAN MEDICINE OPHTHALMOLOGY & OTORHINOLARYNGOLOGY & DERMATOLOGY | 418 |
| JOURNAL OF THE KOREAN LIBRARY AND INFORMATION SCIENCE SOCIETY | 477 |
| JOURNAL OF THE KOREA SOCIETY OF COMPUTER AND INFORMATION | 1912 |
| THE JOURNAL OF ORIENTAL OBSTETRICS & GYNECOLOGY | 404 |
| THE TRANSACTIONS OF THE KOREAN INSTITUTE OF ELECTRICAL ENGINEERS | 787 |
| JOURNAL OF LIFE SCIENCE | 1203 |
| ETRI JOURNAL | 1100 |
| KIPS TRANSACTIONS ON SOFTWARE AND DATA ENGINEERING | 523 |
| JOURNAL OF KOREAN NAVIGATION AND PORT RESEARCH | 463 |
| JOURNAL OF KOREA MULTIMEDIA SOCIETY | 939 |
| SUB TOTAL | 51676 |
| TEST SET | |
| THE JOURNAL OF KOREAN ACADEMY OF PROSTHODONTICS | 444 |
| KOREAN JOURNAL OF FOOD SCIENCE AND TECHNOLOGY | 869 |
| NUCLEAR ENGINEERING AND TECHNOLOGY | 755 |
| BMB REPORTS | 647 |
| JOURNAL OF KOREAN SOCIETY OF DENTAL HYGIENE | 873 |
| JOURNAL OF KOREA ACADEMIA-INDUSTRIAL COOPERATION SOCIETY | 7109 |
| JOURNAL OF BROADCAST ENGINEERING | 667 |
| JOURNAL OF THE KOREA SOCIETY INDUSTRIAL INFORMATION SYSTEM | 568 |
| JOURNAL OF IKEEE | 779 |
| JOURNAL OF CONVERGENCE FOR INFORMATION TECHNOLOGY | 620 |
| SUB TOTAL | 13331 |
| TOTAL | 65007 |